\documentclass[times,twocolumn,final]{elsarticle}

\usepackage{medima}
\usepackage{framed,multirow}
\usepackage{amssymb}
\usepackage{latexsym}
\usepackage{url}
\usepackage[table]{xcolor}
\usepackage{bm}
\usepackage{bbm}
\usepackage{amsmath}
\usepackage{stackengine}
\usepackage{scalerel}
\usepackage{enumitem}
\usepackage{booktabs}
\usepackage{graphicx}
\usepackage{caption}
\usepackage{subcaption}
\usepackage{adjustbox}
\usepackage{pifont}
\usepackage{hyperref}

\definecolor{newcolor}{rgb}{.8,.349,.1}
\journal{Medical Image Analysis}

\begin{document}
\verso{X. Wang \textit{et~al.}}
\begin{frontmatter}

\title{PointExplainer: Towards Transparent Parkinson's Disease Diagnosis}

\author[1]{Xuechao Wang\corref{cor1}}
\cortext[cor1]{Corresponding author:}
\ead{xuechao.wang@ugent.be}
\author[2]{Sven N{\~o}mm}
\author[1]{Junqing Huang}
\author[3]{Kadri Medijainen}
\author[4]{Aaro Toomela} 
\author[1,5]{Michael Ruzhansky}

\address[1]{Department of Mathematics: Analysis, Logic and Discrete Mathematics, Ghent University, Ghent, Belgium}
\address[2]{Department of Software Science, Faculty of Information Technologies, Tallinn University of Technology, Akadeemia tee 15a, 12618, Tallinn, Estonia}
\address[3]{Institute of Sport Sciences and Physiotherapy, University of Tartu, Puusepa 8, Tartu 51014, Estonia}
\address[4]{School of Natural Sciences and Health, Tallinn University, Narva mnt. 25, 10120, Tallinn, Estonia}
\address[5]{School of Mathematical Sciences, Queen Mary University of London, Mile End Road, London E1 4NS, United Kingdom}

\received{XXX}
\finalform{XXX}
\accepted{XXX}
\availableonline{XXX}

\begin{abstract}
Deep neural networks have shown potential in analyzing digitized hand-drawn signals for early diagnosis of Parkinson's disease. However, the lack of clear interpretability in existing diagnostic methods presents a challenge to clinical trust. In this paper, we propose PointExplainer, an explainable diagnostic strategy to identify hand-drawn regions that drive model diagnosis. Specifically, PointExplainer assigns discrete attribution values to hand-drawn segments, explicitly quantifying their relative contributions to the model's decision. Its key components include: (i) a diagnosis module, which encodes hand-drawn signals into 3D point clouds to represent hand-drawn trajectories, and (ii) an explanation module, which trains an interpretable surrogate model to approximate the local behavior of the black-box diagnostic model. We also introduce consistency measures to further address the issue of faithfulness in explanations. Extensive experiments on two benchmark datasets and a newly constructed dataset show that PointExplainer can provide intuitive explanations with no diagnostic performance degradation. The source code is available at \url{https://github.com/chaoxuewang/PointExplainer}.
\end{abstract}

\end{frontmatter}

\section{Introduction}\label{sec:introduction}
Parkinson's disease (PD) is one of the most prevalent neurological disorders worldwide, leading to a decrease in functional, cognitive, and behavioral abilities \cite{aarsland2021parkinson, bloem2021parkinson}. Despite the unclear etiology and lack of a cure, evidence indicates that early diagnosis, coupled with subsequent neuroprotective interventions, can significantly delay its progression \cite{salat2016challenges}. Hand drawing is a common but complex human activity, requiring fine motor control and involving a sophisticated interplay of cognitive, sensory, and perceptual-motor functions \cite{carmeli2003aging}. Dysgraphia is recognized as a crucial biomarker in the early stages of PD \cite{letanneux2014micrographia}. 

Digitized hand-drawn analysis \cite{aouraghe2022literature, guo2022tree}, as a noninvasive and easily accessible biometric technology, has emerged as a promising computer-aided approach for diagnosing PD \cite{drotar2014analysis,impedovo2019velocity,diaz2021sequence,wang2024lstm,pereira2016deep,diaz2019dynamically}. This approach uses a digital tablet \cite{isenkul2014improved} or smart pen \cite{barth2012combined}, in combination with specific hand-drawn templates, to capture biological time-series signals generated during the subject's hand-drawing process \cite{isenkul2014improved,valla2022tremor}. These signals are used to quantify the clinical manifestations of dysgraphia in PD patients, which are difficult to objectively assess with traditional pen-and-paper methods \cite{alty2017use}. Combined with machine learning technology, it has achieved diagnostic performance comparable to that of experts \cite{thomas2017handwriting,mei2021machine}.

However, it is still challenging to understand the model's diagnostic logic, as the high heterogeneity of PD \cite{tolosa2021challenges} and the black-box nature of diagnostic models \cite{ribeiro2016should}. In the medical field, the lack of transparency is particularly concerning \cite{billot2025spatial,fraser2018need,scorza2021surgical}. By presenting clear (textual or visual) explanations, clinicians can utilize their expertise to decide whether to trust the model’s predictions or evaluate whether the model performs as anticipated, aiding in making more accurate diagnoses \cite{xia2025interpretable}. To alleviate this problem, an ideal potential diagnostic paradigm \cite{tjoa2020survey} could involve introducing an interpretable surrogate model (i.e., explainer) to reveal the behavior of the black-box diagnostic model \cite{simonyan2013deep,jha2018disentangling,singla2023explaining}. 

\begin{figure}[t]
  \centering  
  \vspace{-1mm}
  \includegraphics[width=0.4\textwidth]{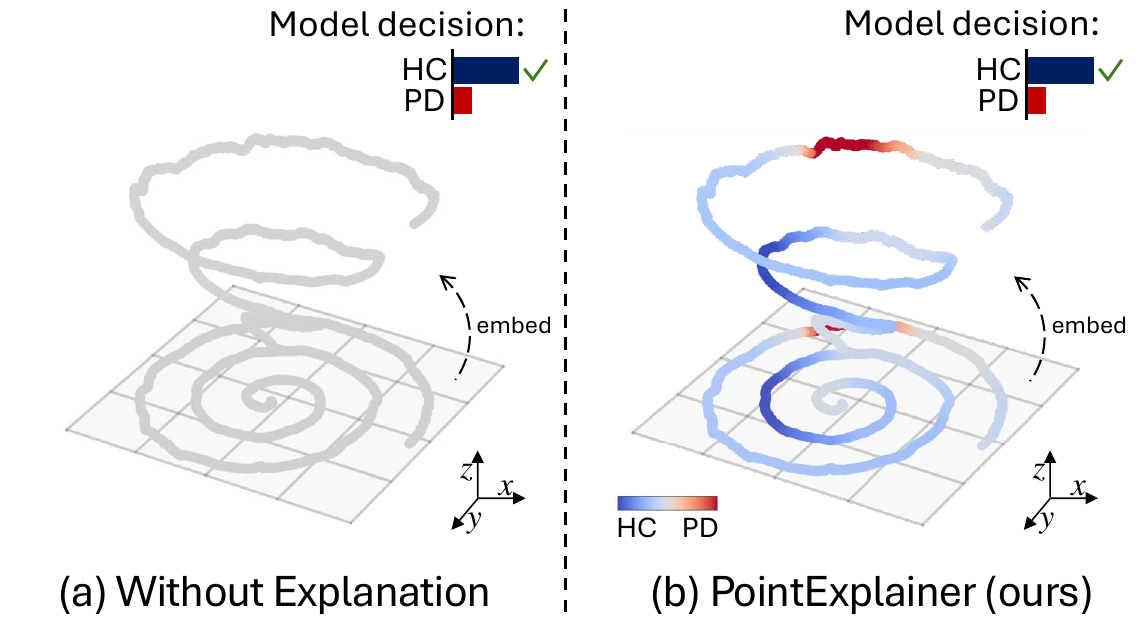}
  \vspace{-1mm}
  \caption{Visualization of diagnostic results for the same instance. The 2D spiral hand-drawn trajectory embeds an additional hand-drawn feature (e.g., radius, which is the distance of the hand-drawn point relative to the template center.) as height point by point to form a point cloud. (a) Most methods only output the final model decision, whereas (b) our PointExplainer enhances interpretability by generating an attribution map. This map overlays attribution values as color intensities onto the hand-drawn trajectory, intuitively highlighting key regions. This makes the model decision more transparent. \protect\textcolor{blue}{Blue} represents a tendency towards healthy controls (HC), and \protect\textcolor{red}{red} represents a tendency towards Parkinson’s disease (PD)\protect\footnotemark.}
  \label{fig:1}
  \vspace{-3mm}
\end{figure}

Following previous work in \cite{wang2024lstm}, we observed that the comprehensive diagnostic results can be determined by analyzing local hand-drawn regions. However, this method fails to fully consider the potential interactions between hand-drawn regions and ignores their differences in diagnostic significance. To this end, further explicitly assigning relative importance values to hand-drawn segments becomes a natural and logical choice. In this paper, we propose a more intuitive diagnostic strategy, PointExplainer, the core of which is to provide a discrete attribution map as the explanation of hand drawing inputs. As shown in Fig.\ref{fig:1}, PointExplainer presents an intelligible representation of the relationship between hand-drawn points and model decisions. This enables clinicians to quantitatively and qualitatively understand which hand-drawn segments are crucial to the model's diagnosis. In general, our contributions could be summarized as follows:
\begin{itemize}
    \item We introduce an intuitive perspective to assigning attribution values to hand-drawn segments, enabling clinicians to easily localize diagnostically relevant regions and obtain clear explanations aligned with human reasoning.  
    \item We propose an explainable diagnostic strategy, PointExplainer, which precisely encodes hand-drawn trajectories as point clouds and introduces an explainer to intelligibly present the decision logic of the black-box diagnostic model.
    \item We design two consistency measurement methods with four evaluation metrics to quantitatively verify the local consistency of model behavior, ensuring the faithfulness of explanations.
    \item Extensive experiments and analyses on diverse datasets demonstrate the effectiveness and robustness of our method, enhancing the practicality and applicability of digital hand-drawing analysis for PD diagnosis.
\end{itemize}

\section{Related Work}\label{sec:related work}

\subsection{Parkinson's Disease Diagnosis.} \label{sec:pd diagnosis}
Recent studies have extensively explored the quantitative analysis of hand-drawn impairments in PD \cite{aouraghe2022literature}. A feasible approach is to manually design discriminative features from hand-drawn signals for machine learning models \cite{drotar2014analysis,impedovo2019velocity}. However, this approach relies on handcrafted features, limiting adaptability to individual variations and potentially undermining the model's effectiveness. To alleviate this, an alternative approach is to explore the integration of deep neural networks (DNNs) to leverage their powerful feature extraction capabilities \cite{sigcha2023deep}. Currently, recurrent neural networks (RNNs) \cite{lipton2015critical} have shown potential to capture time-dependent patterns in hand-drawn signals \cite{diaz2021sequence,wang2024lstm}. In addition, some studies transform hand-drawn signals into images, reframing the PD diagnosis task as an image recognition problem \cite{pereira2016deep,diaz2019dynamically}, achieving breakthroughs in spatial feature extraction. However, image-based representation methods may fall short of fully capturing the complexity of hand-drawn trajectories, as pixel-grid representations are prone to lose fine-grained motion details during discretization, posing challenges for high-precision analysis of subtle motor changes in PD. To overcome this, we extend this concept, embedding the hand-drawn signal points by points into a point cloud. This method not only faithfully reproduces the geometric structure of hand-drawn trajectories but also effectively integrates the collected independent hand-drawn features.

\footnotetext{A slightly moving window smoothing is used to reduce abrupt color changes at segment boundaries.}
\begin{figure*}[t]
  \centering  
  \vspace{-1mm}
  \includegraphics[width=1.0\textwidth]{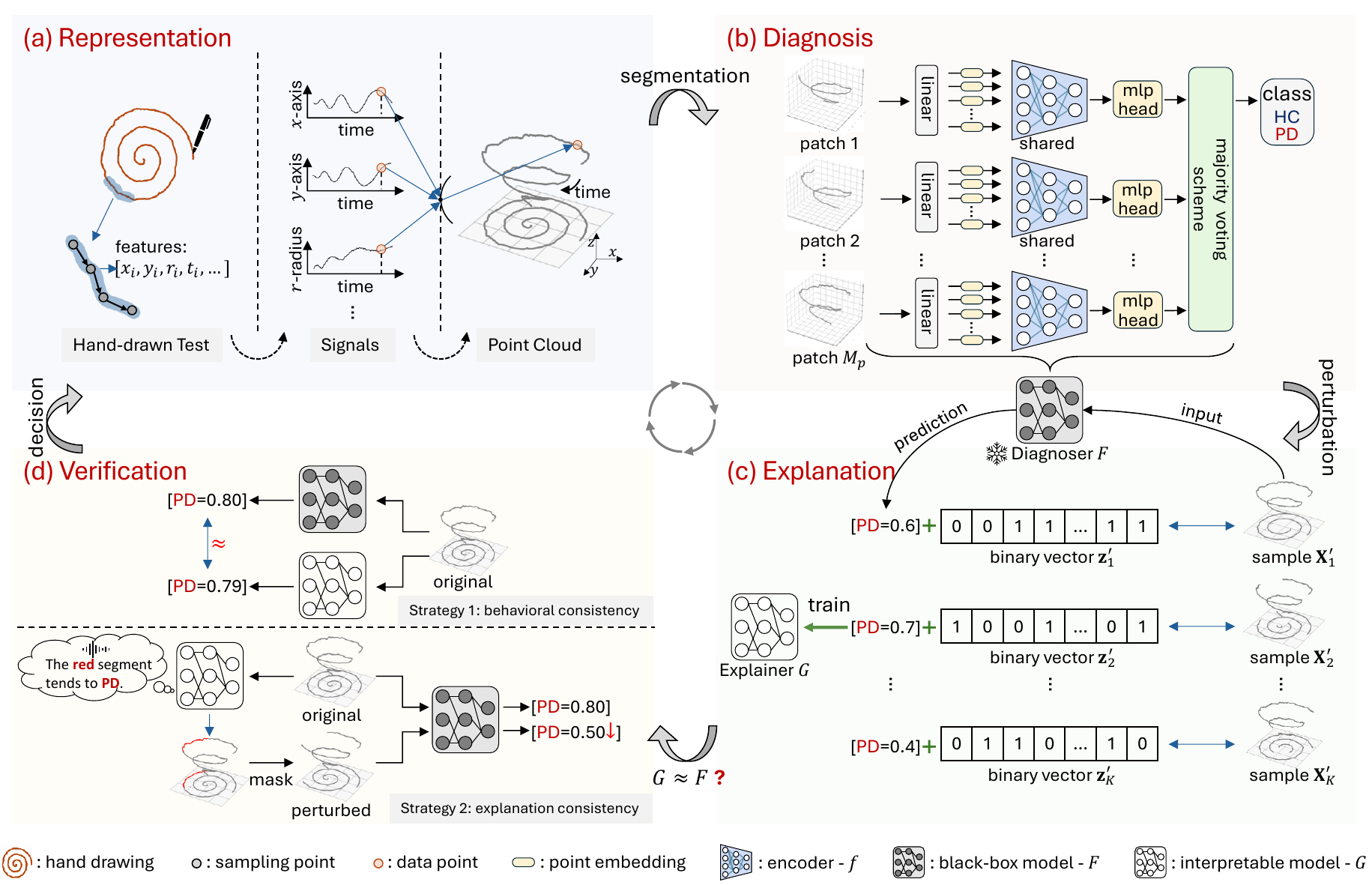}
  \vspace{-1mm}
  \caption{Overview of the proposed PointExplainer framework, consisting of four main modules: (a) point cloud representation, (b) black-box diagnosis, (c) point perturbation explanation, and (d) explanation fidelity verification. First, the hand-drawn signal is encoded point by point into a point cloud. The black-box diagnostic model then classifies the constructed point cloud representation. Next, an interpretable model identifies key hand-drawn regions, and finally, the reliability of explanations is quantitatively verified. Further details can be found in Section \ref{sec:method}.}
  \label{fig:2}
  \vspace{-3mm}
\end{figure*}

\subsection{Point-Cloud Recognition Models.} \label{sec:pointcloud model}
Deep learning-based approaches for recognizing 3D point clouds can be categorized into the following three main types: projection-based, voxel-based, and point-based \cite{guo2020deep}. The multi-view projection approaches \cite{chen2017multi,kanezaki2018rotationnet} adopt projecting 3D point clouds into various 2D image planes, followed by extracting feature representations using 2D CNNs from these projections. A fusion process is then conducted to integrate the multi-view features and form the final output representations. However, the network performance may be heavily affected by the choice of projection planes and occlusion in 3D. Another approach involves 3D voxelization, which generally rasterizes point clouds onto regular grids and applies 3D convolution for feature learning \cite{maturana2015voxnet,song2017semantic,wang2024comparison}. However, this strategy can result in substantial computational and memory expenses due to the cubic increase in voxel numbers with resolution. In addition, these techniques may sacrifice geometric detail due to quantization onto the voxel grid. Recently, rather than projecting or quantizing irregular point clouds onto regular grids in 2D or 3D, some works \cite{qi2017pointnet,qi2017pointnet++,shen2018mining,wu2019pointconv,zhao2021point} have developed deep network architectures that directly ingest point clouds, as sets embedded in continuous space. This strategy not only preserves more geometric details and local structural information but also avoids the computational and memory burdens associated with voxelization. In this context, we use a vanilla point-based network as the point-cloud recognition model.

\subsection{Explainable Artificial Intelligence.} \label{sec:xai}
Explainable artificial intelligence (XAI) aims to provide explanations that facilitate users' understanding of model decisions \cite{tjoa2020survey,arrieta2020explainable,billot2025spatial}. Two research topics in XAI offer two perspectives on explainability: built-in explainability and post-hoc explainability. As the names imply, built-in explainability approaches \cite{chen2019looks,alvarez2018towards} inherently provide built-in explanations generated by the model itself. The prevailing strategy is to use directly interpretable models, e.g., decision trees, rules \cite{wang2015falling}, additive models \cite{caruana2015intelligible}, attention-based networks \cite{selvaraju2017grad}, or sparse linear models \cite{ustun2016supersparse}. However, this approach suffers from a trade-off between model flexibility and explainability  \cite{tjoa2020survey}, implying that one cannot use a model with very complex behavior, but expect humans to fully comprehend it. In contrast, post-hoc explainability approaches \cite{simonyan2013deep,jha2018disentangling} aim to separate explanations from models, i.e., extracting post-hoc explanations by treating the original model as a black box. This strategy allows the model to be as flexible and accurate as possible, e.g., attribution-based methods \cite{sundararajan2017axiomatic,selvaraju2017grad}, local perturbation-based methods \cite{zeiler2014visualizing}, and concept-based methods \cite{koh2020concept}. In this paper, consistent with the post-hoc approach, we introduce an interpretable surrogate model (i.e., explainer) to approximate the behavior of the black-box diagnostic model at the local scope of instances of interest. In addition, we provide objective quantitative validation rather than the subjective interpretation typically used in XAI evaluations \cite{agarwal2023evaluating}. This enables a more rigorous assessment of explanation reliability, ensuring that our method not only provides interpretability but also aligns with measurable evaluation criteria.


\section{Method}\label{sec:method}
In this section, we introduce the proposed PointExplainer. Fig.\ref{fig:2} shows an overview of our method, which builds upon the standard post-hoc explainability framework \cite{tjoa2020survey}. Section \ref{sec:preliminaries} briefly introduces the basic notations and definitions. Section \ref{sec:modelling} shows how to represent hand-drawn signals in the form of point clouds. Section \ref{sec:diagnosis} describes the black-box diagnostic process. Section \ref{sec:explain} trains an interpretable surrogate model. Section \ref{sec:verify} objectively verifies the reliability of explanations.

\subsection{Preliminaries} \label{sec:preliminaries}

\subsubsection{Point Cloud.} 
A point cloud \cite{qi2017pointnet} is represented as $\mathbf{X} \triangleq \{ \mathbf{x}_i \}_{i=1,\ldots,N}$, where $\mathbf{x}_i \in \mathbb{R}^3$ is a 3D point, and $N$ is the number of points in the point cloud. Here, we define a superpoint $\mathbf{S} \triangleq \{ \mathbf{x}_{i_{s}} \}_{i_{s}=1,\ldots,N_{s}}$ as a cluster of points (similar to a word in text or a superpixel in images), where $\mathbf{S} \subseteq \mathbf{X}$, and $N_{s}$ is the number of points contained in the superpoint. In this paper, the grouping of a superpoint is based on the natural sequence of the hand-drawn trajectory, i.e., a superpoint essentially represents a continuous segment of the hand drawing, where the points within share similar semantics \cite{qi2017pointnet++}.

\subsubsection{Point Perturbation.} \label{sec:perturnation}
\begin{figure}[t]
  \centering  
  \vspace{-1mm}
  \includegraphics[width=0.4\textwidth]{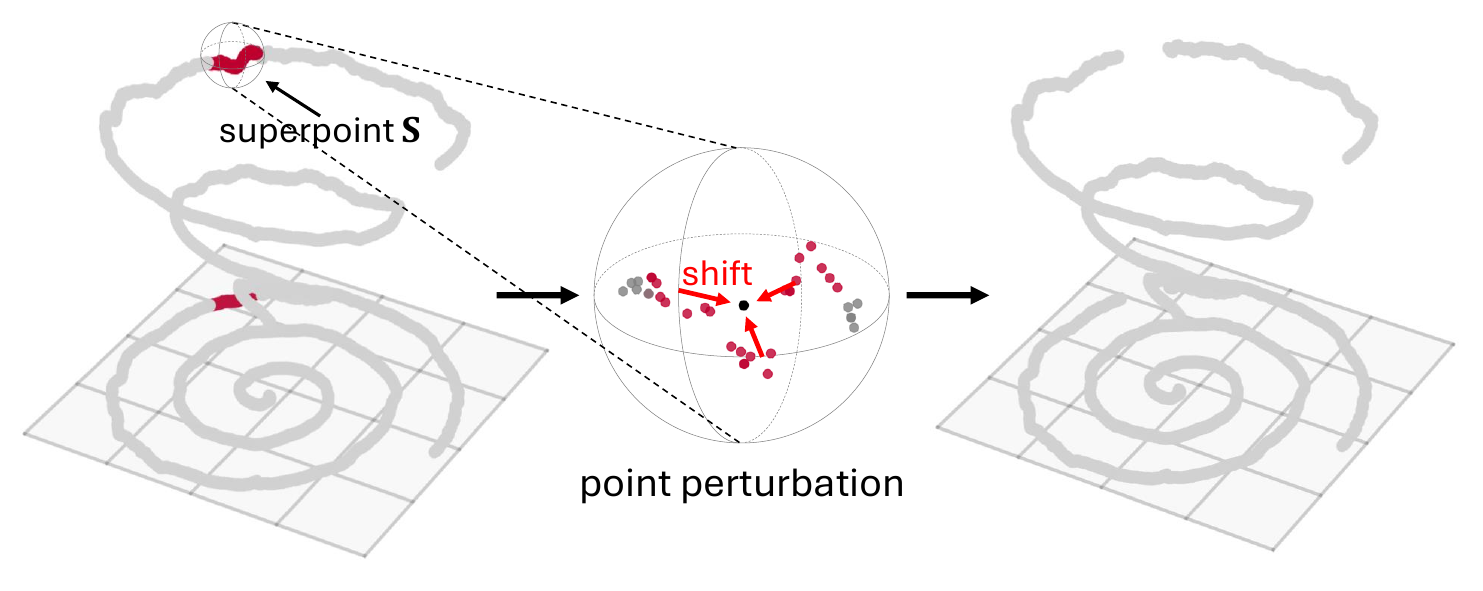}
  \vspace{-1mm}
  \caption{Illustration of Point Perturbation. All points within the superpoint $\mathbf{S}$ are shifted toward its center, eliminating the original local spatial semantic structure.}
  \label{fig:3}
  \vspace{-3mm}
\end{figure}

To eliminate the semantic information contained in a superpoint $\mathbf{S}$, we use a differentiable operation to shift all points within the superpoint $\mathbf{S}$ to its center, as shown in Fig.\ref{fig:3}. This makes intuitive sense, as all the points will occupy the same position after the coordinate translation, making the entire set of points uninformative for classification \cite{zheng2019pointcloud}. Under this operation, the spatial distribution that initially provided semantic meaning is effectively neutralized. Consequently, the contribution of the superpoint $\mathbf{S}$ to recognition can be neglected. 

\subsubsection{Point Explanation.}\label{sec:representation}
An essential criterion for explanations is that they must consider the limitations of the target audience, i.e., intelligible explanations need to use a representation that is easily understandable to users, regardless of the features actually used by the model \cite{jin2023guidelines,arrieta2020explainable}. For example, machine learning practitioners may be able to explain intricate handcrafted features, but clinicians may prefer an explanation that involves a small number of weighted features \cite{subramaniam2024grand}. To align more closely with the clinical diagnosis, we use a set of fix-sized superpoints $\{ \mathbf{S}_{j_{s}}\}_{j_{s}=1,\ldots,M_{s}}$ as the interpretable data representation of a point cloud $\mathbf{X}$, and endeavor to assign relative attribution values to them \cite{lundberg2017unified}. These superpoints satisfy $\bigcap_{j_{s}=1}^{M_{s}} \mathbf{S}_{j_{s}} = \emptyset$ and $\bigcup_{j_{s}=1}^{M_{s}} \mathbf{S}_{j_{s}} = \mathbf{X}$, where $M_{s}$ represents the number of superpoints.

\subsection{Point Cloud Representation}\label{sec:modelling}
Hand drawings possess inherent spatial characteristics, e.g., coordinate features $\left(x,y \right)$. A point cloud is essentially a set of points embedded in space \cite{zhao2021point}, capable of reconstructing the hand-drawn trajectory point by point while precisely preserving subtle movement details that reflect the motor symptoms of PD \cite{chen2021shape}. Therefore, modeling hand-drawn signals as point clouds is intuitively valid and natural. 

To this end, we flesh out this intuition and propose to represent hand-drawn signals in the form of point clouds. As shown in Fig.\ref{fig:2}(a), we first encode the $2$D geometric coordinates of the data points using the hand-drawn features $\left(x,y \right)$. Next, each data point is embedded with an additional hand-drawn feature as its height attribute $z$, e.g., the radius, which is the distance of the sampling point relative to the template center\footnote{We use the first sampling point of the hand-drawn signal as the template center.}. This allows us to construct a 3D point cloud $\mathbf{X}$ within a spherical coordinate system. Furthermore, additional attributes (e.g., color) could be attached by computing velocity or other derived features. Notably, even in the basic setting, where each data point is represented by only three attributes $\left(x,y,z \right)$, our representation method still shows competitive diagnostic performance in experiments.

\subsection{Black-Box Diagnosis}\label{sec:diagnosis}
Next, we elaborate on how recognition is performed using the constructed point cloud $\mathbf{X}$. In hand-drawing tasks, patients with PD are more prone to motor impairments, such as tremors and rigidity \cite{tolosa2021challenges}. These impairments typically manifest as localized abnormalities in hand-drawn trajectories and are reflected as region-specific variations in the point cloud representation.

In view of this, we use a patch-based local analysis strategy \cite{wang2024lstm}. As shown in Fig.\ref{fig:2}(b), the point cloud $\mathbf{X}$ is first segmented into a set of fixed-size patches $\{\mathbf{P}_{j_{p}}\}_{j_{p}=1,...,M_{p}}$ using the sliding window approach \cite{wang2024lstm}, where $M_{p}$ represents the number of patches. Each patch $\mathbf{P}_{j_{p}}$ consists of $w$ consecutive points, sampled according to their drawing order, denoted as $\mathbf{P}_{j_{p}}\triangleq \{ \mathbf{x}_{i_{p}} \mid i_{p}=j_{p} \cdot s+i, i=1,...,w\}$, where $s$ is the step size controlling the overlap between adjacent patches, and $w$ is the window size determining the number of points within each patch. Next, a vanilla point-cloud recognition model, denoted as $f$, processes each patch $\mathbf{P}_{j_{p}}$ as a 1D sequence of point embeddings\footnote{Normalization: the ($x,y$) coordinates are scaled to a unit circle, and other attributes are standardized to mean 0 and variance 1.} and outputs the corresponding prediction probability $\hat{y}_{j_{p}}=f(\mathbf{P}_{j_{p}})$, where $\hat{y}_{j_{p}}$ represents the probability of the patch $\mathbf{P}_{j_{p}}$ being classified as PD. Finally, we use a threshold-based majority voting scheme to determine the final classification result. Given a threshold $\alpha \in [0,1]$, the model decision is computed as $\hat{y}_{final}= \text{vote}(\{\hat{y}_{j_{p}}\}_{j_{p}=1,...,M_{p}}; \alpha)$, where $\hat{y}_{final}$ is the proportion of patches classified as PD within the patch set $\{\mathbf{P}_{j_{p}}\}_{j_{p}=1,...,M_{p}}$. Note that a patch $\mathbf{P}_{j_{p}}$ is classified as PD if $\hat{y}_{j_{p}} \geq \alpha$; otherwise, it is classified as HC. 

Overall, the black-box diagnostic process, denoted as $F$, is defined as $\hat{y}_{final} = F(\mathbf{X}; \alpha)$. Unless otherwise stated, we set the threshold $\alpha=0.5$ to align with standard voting schemes throughout our experiments. We evaluate PointExplainer using a basic recognition model, PointNet, as $f$ to demonstrate the effectiveness of our method. For further details on this model, we refer the readers to \cite{qi2017pointnet}.

\subsection{Point Perturbation Explanation}\label{sec:explain}
Now, we explain the decision logic of the black-box diagnostic model $F$. As shown in Fig.\ref{fig:2}(c), PointExplainer trains an inherently interpretable surrogate model, denoted as $G$, to approximate the local behavior of model $F$. We train the surrogate model $G$ based on the learning framework proposed by \cite{ribeiro2016should}, with adjustments to accommodate point cloud data.

Formally, given an instance $\mathbf{X}=\{ \mathbf{S}_{j_{s}}\}_{j_{s}=1,\ldots,M_{s}}$ to be explained, each superpoint $\mathbf{S}_{j_{s}}$ corresponds to a feature in the surrogate model $G$. The importance of features (e.g., SHAP \cite{lundberg2017unified} value) learned in $G$ represents the relative contribution of superpoints to the final model decision, i.e., their attribution values, denoted as $\{w_{j_{s}}\}_{j_{s}=1,\ldots,M_{s}}$. More specifically, we first generate a set of perturbed samples $\{\mathbf{X}'_{i}\}_{i=1,...,K}$ by randomly perturbing each superpoint (as described in Section \ref{sec:perturnation}), where $K$ is the number of perturbed samples. Note that $\mathbf{X}$ is not included in $\{\mathbf{X}'_{i}\}_{i=1,...,K}$. Then, for each perturbed sample $\mathbf{X}'_{i}$, we construct a corresponding binary vector $\mathbf{z}'_{i}=(z'_{i,j_{s}} \mid j_{s}=1,...,M_{s})$, where each element $z'_{i,j_{s}}\in\{0,1\}$ encodes the perturbation state of the superpoint $\mathbf{S}_{j_{s}}$ in $\mathbf{X}'_{i}$. Here, $z'_{i,j_{s}}=0$ represents that $\mathbf{S}_{j_{s}}$ is in a ``perturbed'' state, while $z'_{i,j_{s}}=1$ indicates that it remains ``unperturbed''. Next, we obtain the counterpart $\hat{y}'_{i,final}$ of the binary vector $\mathbf{z}'_{i}$ from the black-box diagnostic model $F$, where $\hat{y}'_{i,final}=F(\mathbf{X}'_{i}; \alpha)$. Finally, we train the interpretable surrogate model $G$ using the paired dataset $\{(\mathbf{z}'_{i},\hat{y}'_{i,final})\}_{i=1,...,K}$. Note that the diagnostic model $F$ operates on point clouds, whereas the surrogate model $G$ is trained on binary vectors. In addition, a dedicated surrogate model $G$ is trained for each instance.

Overall, the intelligible explanation is a small list of superpoints with attribution values. If the attribution value $w_{j_{s}}$ is positive (or large), the corresponding contribution of $\mathbf{S}_{j_{s}}$ is considered positive. As reintroducing $\mathbf{S}_{j_{s}}$ to $\mathbf{X}$ (i.e., changing from ``perturbed'' to ``unperturbed'') will increase the prediction probability, favoring the PD category. Conversely, the contribution of $\mathbf{S}_{j_{s}}$ is considered negative. We use linear (linear regression (LR), ridge regression (Ridge), and elastic net (ElasticNet)) and nonlinear (decision tree regression (DT), random forest regression (RF), and XGBoost regression (XGB)) interpretable models as $G$ to comprehensively evaluate PointExplainer in our experiments.

\subsection{Explanation Reliability Verification}\label{sec:verify}
Finally, our objective is to verify whether the surrogate model $G$ captures the local behavior of the black-box diagnostic model $F$. As shown in Fig.\ref{fig:2}(d), we introduce two consistency measurement methods with four evaluation metrics to quantify the reliability of explanations.

\subsubsection{behavioral consistency.}
Behavioral consistency \cite{queen2024encoding} measures the consistency between the outputs of different models for the same instance. An intuitive way is to compare the similarity between the prediction probability of $F$ and the regression score of $G$, using error metrics or decision coefficients \cite{tan2022surrogate}. In this paper, we introduce two behavioral consistency metrics: Probability Consistency (PC) and Category Alignment (CA), which are formulated as:
\begin{itemize}
    \item  Probability Consistency (PC)
    \begin{equation}
        \text{PC} = \left| F(\mathbf{X}) - G(\mathbf{z}) \right|
    \end{equation}
    \item  Category Alignment (CA)
    \begin{equation}
    \text{CA} = \mathbb{I} \Big( \text{sign}(F(\mathbf{X}) - 0.5) = \text{sign}(G(\mathbf{z}) - 0.5) \Big)
    \end{equation}
\end{itemize}
where $\text{sign}(\cdot)$ is a sign function, and $\mathbb{I}(\cdot)$ is an indicator function that takes the value 1 when the condition inside the parentheses holds, and 0 otherwise. The PC metric quantifies the discrepancy between model-predicted probabilities, and the CA metric measures the consistency of model output categories. Therefore, in the experiments, the surrogate model $G$ that achieves superior alignment with the diagnostic model $F$ will exhibit a lower PC score and a higher CA score.

\subsubsection{explanation consistency} 
Explanation consistency \cite{kalakoti2024improving} measures the alignment between the explanations provided by the surrogate model $G$ and the decisions made by the diagnostic model $F$. The intuitive idea is to mask (i.e., point perturbation) the superpoints with specific attributes (positive or negative) according to the explanation, and then observe whether the decision exhibits the expected changes. In this paper, we introduce two explanation consistency metrics: Attribution Consistency (AC) and Direction Alignment (DA), which are formulated as:
\begin{itemize}
    \item  Attribution Consistency (AC)
    \begin{equation}
    \text{AC} = \rho \left( \left\{ |w_{j_s}| \right\}_{j_s=1}^{M_s}, \left\{ |\Delta F_{j_s}| \right\}_{j_s=1}^{M_s} \right)
    \end{equation}
    \item  Direction Alignment (DA)
    \begin{equation}
    \text{DA} = \frac{1}{M_s} \sum_{j_s=1}^{M_s} \mathbb{I} \left( \text{sign}(w_{j_s}) = \text{sign}(\Delta F_{j_s}) \right)
    \end{equation}
\end{itemize}
where $\rho(\cdot)$ calculates the Pearson’s correlation coefficient, and $\Delta F_{j_s}=F(\mathbf{X})-F(\mathbf{X} \backslash \mathbf{S}_{j_{s}})$ denotes the change in model decision after masking the superpoint $\mathbf{S}_{j_{s}}$. The AC metric evaluates the correlation between attribution values $(|w_{j_s}|)$ and decision differences $(|\Delta F_{j_s}|)$, verifying whether perturbations to superpoints with larger attribution values lead to more significant changes in the model decision. The DA metric checks whether the direction (positive or negative) of the superpoint influence is consistent with the direction of the decision change $(\Delta F_{j_s})$. Therefore, in the experiments, the explanation that achieves superior alignment with the model decision will exhibit higher AC and DA scores.


\begin{table*}[!t]
    \small
    \renewcommand{\arraystretch}{1.0}
    \setlength{\tabcolsep}{1.0mm} 
    \caption{\textbf{Comparison of diagnostic results} on the SST, DST, and DraWritePD datasets. The best results are indicated in bold, and the second-best results for each column are underlined. $\dagger$ denotes the default decision threshold $\alpha=0.5$. $\ddagger$ indicates that the decision threshold ($\alpha$) is optimized for highest performance (see Fig.~\ref{fig:4}). }
    \label{Tab.1} 
    \resizebox{\textwidth}{!}{
    \begin{tabular}{l c c c c c c c c c c c c c c c c}
        \toprule
        \multirow{2}*{Method} & \multirow{2}*{Modality} && \multicolumn{4}{c}{\cellcolor{gray!20}SST} && \multicolumn{4}{c}{\cellcolor{gray!20}DST} && \multicolumn{4}{c}{\cellcolor{gray!20}DraWritePD} \\
        &&& Accuracy & Sensitivity & Specificity & $\text{F}_{1}$-score && Accuracy & Sensitivity & Specificity & $\text{F}_{1}$-score && Accuracy & Sensitivity & Specificity & $\text{F}_{1}$-score \\
        \hline 
        \addlinespace[0.5pt]
        Sarin et al. \cite{sarin2023three} & - && 84.67 & - & - & - && \underline{90.91} & - & - & - && - & - & - & -  \\
        \rowcolor{gray!5}
        Gil-Martín et al. \cite{gil2023signal} & 1D+2D && - & - & - & \underline{93.13} && - & - & - & \underline{91.44} && - & - & - & -  \\
        Cantürk et al. \cite{canturk2021fuzzy} & 2D && 63.00 & \underline{73.00} & 48.00 & 69.00 && 86.00 & 88.00 & \textbf{83.00} & 88.00 && - & - & - & -  \\
        \rowcolor{gray!5}
        Wrobel et al. \cite{wrobel2022diagnosing} & - && \textbf{92.38} & - & - & - && 90.89 & - & - & - && - & - & - & -  \\
        Khatamino et al. \cite{khatamino2018deep} & 2D && 82.00 & - & - & - && \textbf{93.33} & - & - & - && - & - & - & -  \\
        \rowcolor{gray!5}
        Valla et al. \cite{valla2022tremor} & - && - & - & - & - && - & - & - & - && 84.33 & 70.00 & \underline{93.33} & -  \\
        N\~{o}mm et al. \cite{nomm2020deep} & 2D && - & - & - & - && - & - & - & - && \underline{88.23} & \underline{82.32} & - & \textbf{87.30}  \\
        \rowcolor{gray!5}
        Wang et al. \cite{wang2024comparison} & 3D && - & - & - & - && - & - & - & - && 85.38 & \textbf{82.50} & 87.25 & 85.51  \\
        \hline
        \addlinespace[0.5pt]
        \textbf{PointExplainer}$\dagger$ (\textit{Ours}) & 3D && $88.10_{\pm5.73}$ & $\boldsymbol{93.41_{\pm2.42}}$ & \underline{$66.67_{\pm18.86}$} & $92.68_{\pm3.45}$ && $85.62_{\pm3.75}$ & \underline{$89.08_{\pm9.08}$} & \underline{$73.33_{\pm18.86}$} & $90.47_{\pm2.92}$ && $\boldsymbol{90.74_{\pm2.62}}$ & $80.16_{\pm6.25}$ & $\boldsymbol{97.22_{\pm3.93}}$ & \underline{$86.32_{\pm4.23}$} \\
        \rowcolor{gray!5}
        \textbf{PointExplainer}$\ddagger$ (\textit{Ours}) & 3D && \underline{$89.38_{\pm6.85}$} & $\boldsymbol{93.41_{\pm2.42}}$ & $\boldsymbol{73.33_{\pm24.94}}$ & $\boldsymbol{93.46_{\pm4.13}}$ && $89.90_{\pm1.71}$ & $\boldsymbol{94.54_{\pm4.54}}$ & \underline{$73.33_{\pm18.86}$} & $\boldsymbol{93.62_{\pm0.94}}$ && $\boldsymbol{90.74_{\pm2.62}}$ & $80.16_{\pm6.25}$ & $\boldsymbol{97.22_{\pm3.93}}$ & \underline{$86.32_{\pm4.23}$} \\
        \addlinespace[0.5pt]
        \bottomrule
    \end{tabular} }
\end{table*}
\begin{figure*}[t]
  \centering  
  \vspace{1.5mm}
  \includegraphics[width=1.0\textwidth]{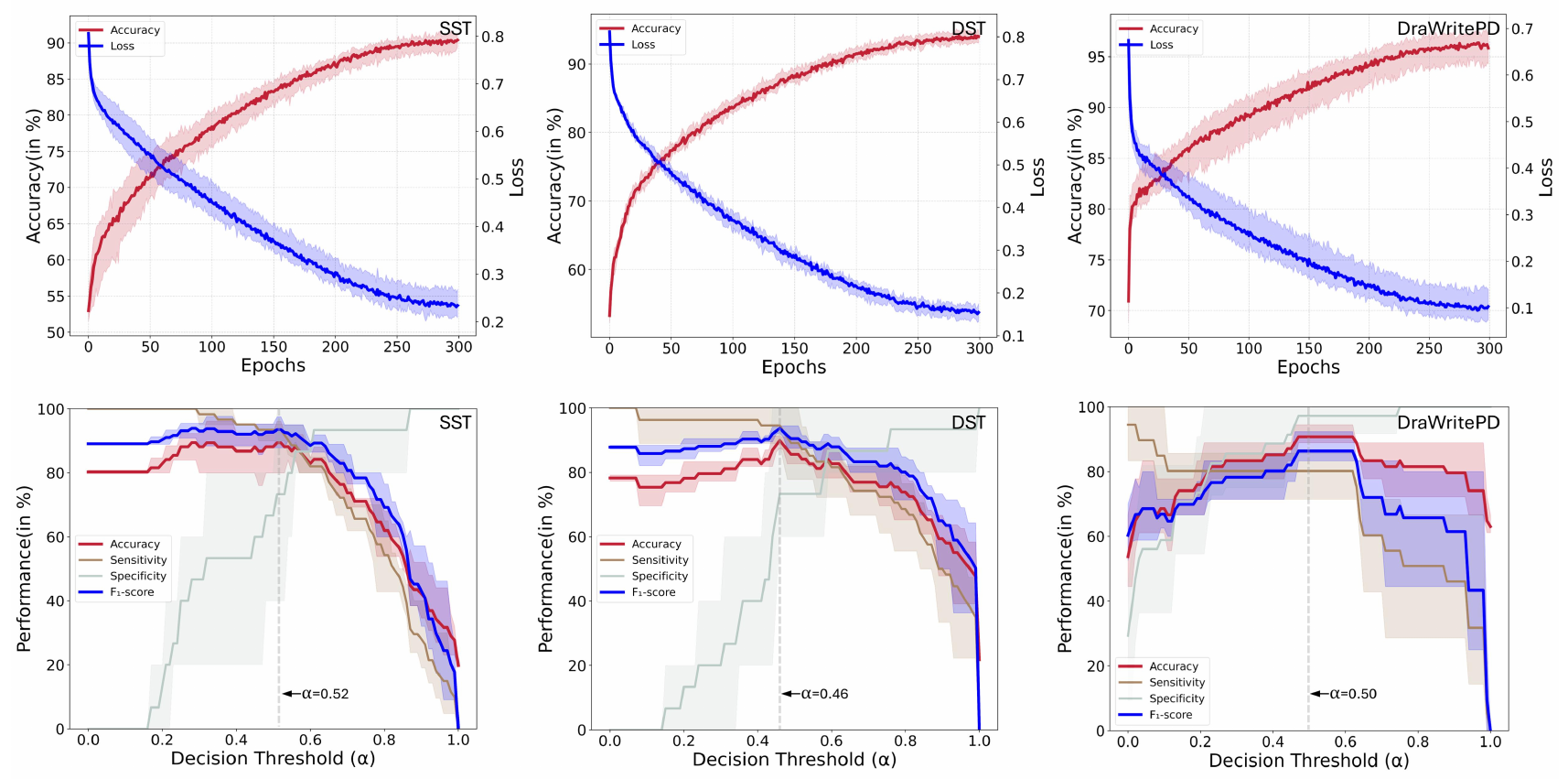}
  \vspace{-1mm}
  \caption{\textbf{Training and performance curves.} We show the training process of PointExplainer and the diagnostic performance under different decision thresholds. The curves represent the mean results from three-fold cross-validation, with shaded regions indicating the range between the minimum and maximum values. The optimal decision thresholds ($\alpha$) for the SST, DST, and DraWritePD datasets are 0.52, 0.46, and 0.50, respectively.}
  \label{fig:4}
\end{figure*}

\section{Experiments}

\subsection{Implementation details} 
\subsubsection{Datasets.}
We conduct experiments on two benchmark datasets: SST \cite{isenkul2014improved}, DST\footnote{\url{https://archive.ics.uci.edu/dataset/395/parkinson+disease+spiral+drawings+using+digitized+graphics+tablet}}, and a newly constructed dataset: DraWritePD \cite{8614244}. We focus on the classic Archimedean spiral task, as shown in Fig.\ref{fig:2}(a). More details are as follows:
\begin{itemize}
    \item SST: $61$ PD patients and $15$ HC subjects. Participants are asked to draw along the lines of a static Archimedean spiral template.
    \item DST: $57$ PD patients and $15$ HC subjects. Participants are asked to draw along the lines of a dynamic Archimedean spiral template, i.e., the template appeared and disappeared intermittently.
    \item DraWritePD: $20$ PD patients and $29$ HC subjects. Participants are asked to draw in the spaces between the static Archimedean spiral lines while avoiding touching the spiral lines.
\end{itemize}
For each sampling point in the acquired hand-drawn signal, the system records six dynamic features, including the x and y coordinates of the pen tip, the pen azimuth and altitude, the pressure exerted on the surface of the tablet by the pen tip, and the timestamp corresponding to each discrete sampling point. In addition, we add three intuitive kinematic features based on this, namely radius, velocity, and acceleration in our experiment.

\begin{table*}[t]
    \small
    \renewcommand{\arraystretch}{1.4}
    \setlength{\tabcolsep}{0.6mm} 
    \caption{\textbf{Benchmarking explanation faithfulness} on the SST, DST, and DraWritePD datasets. Lower Probability Consistency (PC) and higher Category Alignment (CA), Attribution Consistency (AC), and Direction Alignment (DA) indicate superior explanation quality (see Section \ref{sec:verify} for metric definitions). We use both linear and nonlinear interpretable models as explainers for a comprehensive evaluation.}
    \label{Tab.2} 
    \resizebox{\textwidth}{!}{
    \begin{tabular}{l c cccc c cccc c cccc}
        \toprule
        \multirow{2}*{Explainer} & \multirow{2}*{Linear} & \multicolumn{4}{c}{\cellcolor{gray!20}SST} && \multicolumn{4}{c}{\cellcolor{gray!20}DST} && \multicolumn{4}{c}{\cellcolor{gray!20}DraWritePD} \\
        && PC $\downarrow$ & CA $\uparrow$ & AC $\uparrow$ & DA $\uparrow$ && PC $\downarrow$ & CA $\uparrow$ & AC $\uparrow$ & DA $\uparrow$ && PC $\downarrow$ & CA $\uparrow$ & AC $\uparrow$ & DA $\uparrow$ \\
        \hline 
        LR & $\checkmark$ & \underline{$0.0376_{\pm0.0322}$} & \underline{$0.9737_{\pm0.1601}$} & $0.9417_{\pm0.0916}$ &  $0.9769_{\pm0.0427}$ && $0.0530_{\pm0.0451}$ & $0.9130_{\pm0.2818}$ & $0.5683_{\pm0.3568}$ & $0.8644_{\pm0.1451}$ && $0.1182_{\pm0.0779}$ & \underline{$0.9815_{\pm0.1348}$} & $0.4960_{\pm0.3560}$ & $0.8231_{\pm0.1517}$ \\
        \rowcolor{gray!5}
        Ridge & $\checkmark$ & $0.0440_{\pm0.0365}$ & \underline{$0.9737_{\pm0.1601}$} & \underline{$0.9421_{\pm0.0914}$} &  $0.9745_{\pm0.0438}$ && $0.0523_{\pm0.0456}$ & $0.9130_{\pm0.2818}$ & $0.5425_{\pm0.3526}$ & $0.8618_{\pm0.1532}$ && $0.1170_{\pm0.0778}$ & \underline{$0.9815_{\pm0.1348}$} & $0.5288_{\pm0.3559}$ & $0.8246_{\pm0.1472}$ \\
        ElasticNet & $\checkmark$ & $0.0451_{\pm0.0350}$ & \underline{$0.9737_{\pm0.1601}$} & $0.9412_{\pm0.0912}$ & $\boldsymbol{0.9927_{\pm0.0249}}$ && $0.0526_{\pm0.0460}$ & $0.9130_{\pm0.2818}$ & $0.5690_{\pm0.3561}$ & $0.8736_{\pm0.1389}$ && $0.1174_{\pm0.0776}$ & 
        \underline{$0.9815_{\pm0.1348}$} & $0.5272_{\pm0.3561}$ & $0.8419_{\pm0.1450}$ \\
        \rowcolor{gray!5}
        DT & $\times$ & $0.0566_{\pm0.0646}$ & $0.9342_{\pm0.2479}$ & $0.9009_{\pm0.0955}$ &  $0.9768_{\pm0.0402}$ && $0.0303_{\pm0.0418}$ & $0.9130_{\pm0.2818}$ & $0.6699_{\pm0.3126}$ & $0.8776_{\pm0.1440}$ && $0.0829_{\pm0.0807}$ & \underline{$0.9815_{\pm0.1348}$} & \underline{$0.5801_{\pm0.3343}$} &  \underline{$0.8577_{\pm0.1391}$} \\
        RF & $\times$ & $0.0546_{\pm0.0595}$ & $0.9605_{\pm0.1947}$ & $0.9213_{\pm0.0677}$ &  $0.9732_{\pm0.0445}$ &&  \underline{$0.0283_{\pm0.0382}$} & \underline{$0.9275_{\pm0.2593}$} & \underline{$0.7122_{\pm0.2931}$} & \underline{$0.8817_{\pm0.1310}$} && \underline{$0.0820_{\pm0.0753}$} & \underline{$0.9815_{\pm0.1348}$} & $0.5623_{\pm0.3315}$ & $0.8472_{\pm0.1395}$ \\
        \rowcolor{gray!5}
        XGB & $\times$ & $\boldsymbol{0.0204_{\pm0.0220}}$ & $\boldsymbol{0.9868_{\pm0.1140}}$ & $\boldsymbol{0.9579_{\pm0.0423}}$ & \underline{$0.9902_{\pm0.0286}$} && $\boldsymbol{0.0156_{\pm0.0236}}$ & $\boldsymbol{0.9420_{\pm0.2337}}$ & $\boldsymbol{0.7703_{\pm0.2528}}$ & $\boldsymbol{0.9108_{\pm0.1132}}$ && $\boldsymbol{0.0453_{\pm0.0420}}$ & $\boldsymbol{1.0000_{\pm0.0000}}$ & $\boldsymbol{0.6759_{\pm0.3068}}$ & $\boldsymbol{0.8874_{\pm0.1217}}$ \\
        \addlinespace[0.5pt]
        \bottomrule
    \end{tabular} }
\end{table*}

\begin{figure*}[!t]
  \centering  
  \vspace{-1mm}
  \includegraphics[width=1.0\textwidth]{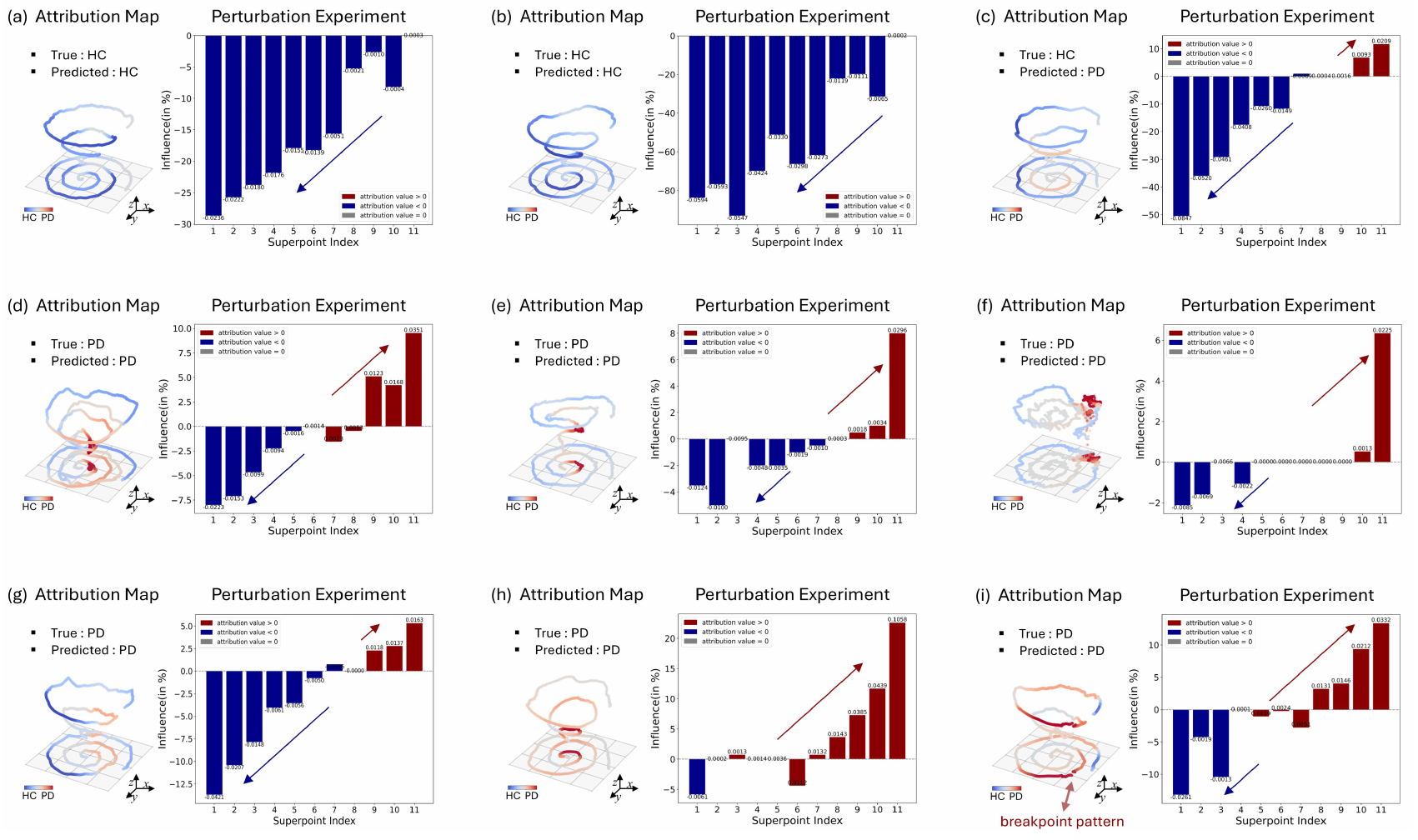}
  \vspace{-1mm}
  \caption{\textbf{Qualitative explanation results and the corresponding perturbation analysis.} We conduct superpoint-level quantitative perturbation experiments based on the explanation results (i.e., attribution maps). For each instance, the left side presents the ground-truth label, model-predicted label, and corresponding attribution map. The right side shows how the model decision changes when each superpoint is perturbed individually (see Section \ref{sec:perturnation} for details on the perturbation strategy). In the attribution map, \textcolor{blue}{blue} indicates attribution values less than 0 (i.e., the superpoint contributes negatively, favoring HC), and \textcolor{red}{red} indicates attribution values greater than 0 (i.e., the superpoint contributes positively, favoring PD). In the bar chart, the x-axis represents the index of superpoints reordered by their attribution values, with the corresponding attribution value labeled on top of each bar, while the y-axis represents the relative percentage change in the predicted probability, computed according to Eq. $(F(\mathbf{X})-F(\mathbf{X} \backslash \mathbf{S}_{j_{s}}))/F(\mathbf{X})$.}
  \label{fig:5}
  \vspace{-3mm}
\end{figure*}

\begin{table*}[t]
    \caption{Ablation studies of \textit{height feature}, \textit{window size} and \textit{attribute importance} on the SST, DST, and DraWritePD datasets. (a-c) Diagnostic differences when using different hand-drawn features as the point cloud height. (d-f) Impact of different window sizes on diagnostic performance. (g-i) Importance of point cloud height and color attributes. \textit{w/o Height} represents the height $z=1$, \textit{w/o Color} indicates the color $(r,g,b)=(1, 1, 1)$. Default settings are marked in\colorbox{orange!15}{orange}.}
    \label{Tab.3}
    \resizebox{\textwidth}{!}{
    \begin{tabular}{ccccc}
        \begin{subtable}{0.4\textwidth}
            \captionsetup{font=normalsize}
            \small
             \renewcommand{\arraystretch}{1.0}
            \setlength{\tabcolsep}{0.5mm} 
            \centering
            \resizebox{\textwidth}{!}{
            \begin{tabular}{lcccc}
                \toprule
                \multirow{2}{*}{Feature} & \multicolumn{4}{c}{\cellcolor{gray!20}SST} \\
                & Accuracy & Sensitivity & Specificity & $\text{F}_{1}$-score  \\
                \hline
                azimuth  & $76.31_{\pm3.29}$ & $72.14_{\pm6.14}$ & $\boldsymbol{93.33_{\pm9.43}}$ & $82.87_{\pm3.16}$ \\
                altitude & $80.36_{\pm6.10}$ & $83.73_{\pm8.12}$ & $66.67_{\pm9.43}$ & $87.07_{\pm4.36}$ \\
                pressure & \cellcolor{orange!15}\underline{$88.10_{\pm5.73}$} & \cellcolor{orange!15}$\boldsymbol{93.41_{\pm2.42}}$ & \cellcolor{orange!15}$66.67_{\pm18.86}$ & \cellcolor{orange!15}$\boldsymbol{92.68_{\pm3.45}}$  \\
                radius & $\boldsymbol{88.21_{\pm5.52}}$ & \underline{$91.98_{\pm8.06}$} & $73.33_{\pm24.94}$ & \underline{$92.50_{\pm3.60}$} \\
                velocity & $78.97_{\pm3.57}$ & $78.81_{\pm5.60}$ & \underline{$80.00_{\pm16.33}$} & $85.67_{\pm2.71}$ \\
                acceleration & $80.36_{\pm5.15}$ & $82.14_{\pm7.85}$ & $73.33_{\pm9.43}$ & $86.85_{\pm4.09}$ \\
                \bottomrule
            \end{tabular}}
            \caption{Height Feature}
            \label{tab:sub1}
        \end{subtable}  &&
        
        \begin{subtable}{0.4\textwidth}
            \captionsetup{font=normalsize}
            \small
            \renewcommand{\arraystretch}{1.0}
            \setlength{\tabcolsep}{0.5mm} 
            \centering
            \resizebox{\textwidth}{!}{
            \begin{tabular}{lcccc}
                \toprule
                \multirow{2}{*}{Feature} & \multicolumn{4}{c}{\cellcolor{gray!20}DST} \\
                & Accuracy & Sensitivity & Specificity & $\text{F}_{1}$-score  \\
                \hline
                 azimuth  & $82.90_{\pm8.68}$ & $81.85_{\pm9.50}$ & $\boldsymbol{86.67_{\pm9.43}}$ & $87.99_{\pm6.57}$ \\
                altitude & \underline{$85.55_{\pm1.59}$} & $\boldsymbol{96.39_{\pm2.55}}$ & $46.67_{\pm9.43}$ & $\boldsymbol{91.26_{\pm0.89}}$ \\
                pressure & $84.16_{\pm3.54}$ & \underline{$92.88_{\pm6.54}$} & $53.33_{\pm9.43}$ & $90.09_{\pm2.55}$ \\
                radius & \cellcolor{orange!15}$\boldsymbol{85.62_{\pm3.75}}$ & \cellcolor{orange!15}$89.08_{\pm9.08}$ & \cellcolor{orange!15}\underline{$73.33_{\pm18.86}$} & \cellcolor{orange!15}\underline{$90.47_{\pm2.92}$} \\
                velocity & $78.11_{\pm4.33}$ & $85.15_{\pm2.75}$ & $53.33_{\pm18.86}$ & $85.91_{\pm2.70}$ \\
                acceleration & $70.98_{\pm2.34}$ & $79.69_{\pm1.94}$ & $40.00_{\pm16.33}$ & $81.14_{\pm0.96}$ \\
                \bottomrule
            \end{tabular}}
            \caption{Height Feature}
            \label{tab:sub2}
        \end{subtable}  &&
        
        \begin{subtable}{0.4\textwidth}
            \captionsetup{font=normalsize}
            \small
            \renewcommand{\arraystretch}{1.0}
            \setlength{\tabcolsep}{0.5mm} 
            \centering
            \resizebox{\textwidth}{!}{
            \begin{tabular}{lcccc}
                \toprule
                \multirow{2}{*}{Feature} & \multicolumn{4}{c}{\cellcolor{gray!20}DraWritePD} \\
                & Accuracy & Sensitivity & Specificity & $\text{F}_{1}$-score  \\
                \hline
                azimuth  & $85.19_{\pm2.62}$ & $70.63_{\pm10.71}$ & \underline{$94.44_{\pm7.86}$} & $77.66_{\pm4.36}$ \\
                altitude & $79.63_{\pm2.62}$ & $55.56_{\pm9.78}$ & $94.19_{\pm4.12}$ & $66.46_{\pm5.20}$ \\
                pressure & $79.63_{\pm5.24}$ & \underline{$74.60_{\pm8.09}$} & $82.07_{\pm13.04}$ & $73.41_{\pm2.63}$ \\
                radius & $85.19_{\pm2.62}$ & $65.08_{\pm5.94}$ & $\boldsymbol{97.22_{\pm3.93}}$ & $76.26_{\pm5.00}$ \\
                velocity & \underline{$88.89_{\pm4.54}$} & \underline{$74.60_{\pm8.09}$} & $\boldsymbol{97.22_{\pm3.93}}$ & \underline{$82.79_{\pm8.00}$} \\
                acceleration & \cellcolor{orange!15}$\boldsymbol{90.74_{\pm2.62}}$ & \cellcolor{orange!15}$\boldsymbol{80.16_{\pm6.25}}$ & \cellcolor{orange!15}$\boldsymbol{97.22_{\pm3.93}}$ & \cellcolor{orange!15}$\boldsymbol{86.32_{\pm4.23}}$ \\
                \bottomrule
            \end{tabular}}
            \caption{Height Feature}
            \label{tab:sub3}
        \end{subtable}  \\\\
        
        \begin{subtable}{0.4\textwidth}
            \captionsetup{font=normalsize}
            \small
            \renewcommand{\arraystretch}{1.0}
            \setlength{\tabcolsep}{0.5mm} 
            \centering
            \resizebox{\textwidth}{!}{
            \begin{tabular}{lccccccccc}
                \toprule
                \multirow{2}{*}{Window} &&&&&& \multicolumn{4}{c}{\cellcolor{gray!20}SST} \\
                &&&&&& Accuracy & Sensitivity & Specificity & $\text{F}_{1}$-score  \\
                \hline
                64 &&&&&& $84.15_{\pm5.77}$ & $80.24_{\pm7.25}$ & $\boldsymbol{100.0_{\pm0.0}}$ & $88.85_{\pm4.60}$ \\
                128 &&&&&& \underline{$86.77_{\pm7.62}$} & $83.49_{\pm9.54}$ & $\boldsymbol{100.0_{\pm0.0}}$ & $90.70_{\pm5.90}$ \\
                256 &&&&&& \cellcolor{orange!15}$\boldsymbol{88.10_{\pm5.73}}$ & \cellcolor{orange!15}$93.41_{\pm2.42}$ & \cellcolor{orange!15}\underline{$66.67_{\pm18.86}$} & \cellcolor{orange!15}$\boldsymbol{92.68_{\pm3.45}}$ \\
                512 &&&&&& $85.15_{\pm1.74}$ & $\boldsymbol{96.66_{\pm2.37}}$ & $40.00_{\pm0.00}$ & \underline{$91.20_{\pm1.11}$} \\
                768 &&&&&& $77.87_{\pm3.38}$ & $89.55_{\pm4.19}$ & $33.33_{\pm9.43}$ & $86.47_{\pm2.21}$ \\
                1024 &&&&&& $77.75_{\pm2.07}$ & \underline{$94.26_{\pm0.69}$} & $20.00_{\pm0.00}$ & $86.80_{\pm1.46}$ \\
                \bottomrule
            \end{tabular}}
            \caption{Window Size}
            \label{tab:sub4}
        \end{subtable} &&
        
        \begin{subtable}{0.4\textwidth}
            \captionsetup{font=normalsize}
            \small
            \renewcommand{\arraystretch}{1.0}
            \setlength{\tabcolsep}{0.5mm} 
            \centering
            \resizebox{\textwidth}{!}{
            \begin{tabular}{lccccccccc}
                \toprule
               \multirow{2}{*}{Window} &&&&&& \multicolumn{4}{c}{\cellcolor{gray!20}DST} \\
                &&&&&& Accuracy & Sensitivity & Specificity & $\text{F}_{1}$-score  \\
                \hline
                64 &&&&&& $83.25_{\pm3.81}$ & $82.32_{\pm9.34}$ & \underline{$86.67_{\pm18.86}$} & $88.36_{\pm3.46}$ \\
                128 &&&&&& $84.64_{\pm5.49}$ & $84.07_{\pm11.70}$ & \underline{$86.67_{\pm18.86}$} & $89.25_{\pm4.55}$ \\
                256 &&&&&& \underline{$84.70_{\pm2.06}$} & $82.42_{\pm5.07}$ & $\boldsymbol{93.33_{\pm9.43}}$ & \underline{$89.43_{\pm1.83}$} \\
                512 &&&&&& \cellcolor{orange!15}$\boldsymbol{85.62_{\pm3.75}}$ & \cellcolor{orange!15}$\boldsymbol{89.08_{\pm9.08}}$ & \cellcolor{orange!15}$73.33_{\pm18.86}$ & \cellcolor{orange!15}$\boldsymbol{90.47_{\pm2.92}}$ \\
                768 &&&&&& $81.40_{\pm2.77}$ & \underline{$86.20_{\pm6.42}$} & $66.67_{\pm9.43}$ & $87.56_{\pm2.24}$ \\
                1024 &&&&&& $77.47_{\pm2.98}$ & $76.83_{\pm1.46}$ & $80.00_{\pm16.33}$ & $83.51_{\pm1.73}$ \\
                \bottomrule
            \end{tabular}}
            \caption{Window Size}
            \label{tab:sub5}
        \end{subtable} &&
        
        \begin{subtable}{0.4\textwidth}
            \captionsetup{font=normalsize}
            \small
            \renewcommand{\arraystretch}{1.0}
            \setlength{\tabcolsep}{0.5mm} 
            \centering
            \resizebox{\textwidth}{!}{
            \begin{tabular}{lccccccccc}
                \toprule
                \multirow{2}{*}{Window} &&&&&& \multicolumn{4}{c}{\cellcolor{gray!20}DraWritePD} \\
                &&&&&& Accuracy & Sensitivity & Specificity & $\text{F}_{1}$-score  \\
                \hline
                64 &&&&&& $85.19_{\pm5.24}$ & $69.84_{\pm2.24}$ & \underline{$94.44_{\pm7.86}$} & $77.78_{\pm7.86}$ \\
                128 &&&&&& $87.04_{\pm6.93}$ & \underline{$74.60_{\pm8.09}$} & \underline{$94.44_{\pm7.86}$} & $80.77_{\pm10.62}$ \\
                256 &&&&&& \underline{$88.89_{\pm4.54}$} & \underline{$74.60_{\pm8.09}$} & $\boldsymbol{97.22_{\pm3.93}}$ & \underline{$82.79_{\pm8.00}$} \\
                512 &&&&&& \cellcolor{orange!15}$\boldsymbol{90.74_{\pm2.62}}$ & \cellcolor{orange!15}$\boldsymbol{80.16_{\pm6.25}}$ & \cellcolor{orange!15}$\boldsymbol{97.22_{\pm3.93}}$ & \cellcolor{orange!15}$\boldsymbol{86.32_{\pm4.23}}$ \\
                768 &&&&&& $86.71_{\pm3.08}$ & \underline{$74.60_{\pm8.09}$} & $93.94_{\pm4.29}$ & $80.59_{\pm5.65}$ \\
                1024 &&&&&& $86.71_{\pm3.08}$ & \underline{$74.60_{\pm8.09}$} & $93.94_{\pm4.29}$ & $80.59_{\pm5.65}$ \\
                \bottomrule
            \end{tabular}}
            \caption{Window Size}
            \label{tab:sub6}
        \end{subtable} \\\\
        
        \begin{subtable}{0.4\textwidth}
            \captionsetup{font=normalsize}
            \small
            \renewcommand{\arraystretch}{1.1}
            \setlength{\tabcolsep}{0.5mm} 
            \centering
            \resizebox{\textwidth}{!}{
            \begin{tabular}{lcccc}
                \toprule
                \multirow{2}{*}{Attribute} & \multicolumn{4}{c}{\cellcolor{gray!20}SST} \\
                & Accuracy & Sensitivity & Specificity & $\text{F}_{1}$-score  \\
                \hline
                w/o Height & $71.23_{\pm9.71}$ & $69.21_{\pm15.80}$ & $80.00_{\pm16.33}$ & $78.34_{\pm9.90}$ \\
                w/ \hspace{1mm} Height & \cellcolor{orange!15}$88.10_{\pm5.73}$ & \cellcolor{orange!15}$93.41_{\pm2.42}$ & \cellcolor{orange!15}$66.67_{\pm18.86}$ & \cellcolor{orange!15}$92.68_{\pm3.45}$ \\
                \textit{improvement} & \textcolor{red}{$+16.87 \uparrow$} & \textcolor{red}{$+24.2 \uparrow$} & \textcolor{blue}{$-13.33 \downarrow$} & \textcolor{red}{$+14.34 \uparrow$} \\
                \arrayrulecolor{gray} \hline \arrayrulecolor{black}
                w/o Color & $89.38_{\pm6.85}$ & $95.08_{\pm0.11}$ & $66.67_{\pm33.99}$ & $93.64_{\pm3.87}$ \\
                w/ \hspace{1mm} Color & $86.72_{\pm6.87}$ & $96.75_{\pm2.30}$ & $46.67_{\pm37.71}$ & $92.28_{\pm3.87}$ \\
                \textit{improvement} & \textcolor{blue}{$-2.66 \downarrow$} & \textcolor{red}{$+1.67 \uparrow$} & \textcolor{blue}{$-20.00 \downarrow$} & \textcolor{blue}{$-1.36 \downarrow$} \\
                \bottomrule
            \end{tabular}}
            \caption{Attribute Importance}
            \label{tab:sub4}
        \end{subtable} &&
        
        \begin{subtable}{0.4\textwidth}
            \captionsetup{font=normalsize}
            \small
            \renewcommand{\arraystretch}{1.1}
            \setlength{\tabcolsep}{0.5mm} 
            \centering
            \resizebox{\textwidth}{!}{
            \begin{tabular}{lcccc}
                \toprule
                \multirow{2}{*}{Attribute} & \multicolumn{4}{c}{\cellcolor{gray!20}DST} \\
                & Accuracy & Sensitivity & Specificity & $\text{F}_{1}$-score  \\
                \hline
                 w/o Height & $76.91_{\pm3.50}$ & $76.20_{\pm8.98}$ & $80.00_{\pm16.33}$ & $83.53_{\pm3.46}$  \\
                w/ \hspace{1mm} Height & \cellcolor{orange!15}$85.62_{\pm3.75}$ & \cellcolor{orange!15}$89.08_{\pm9.08}$ & \cellcolor{orange!15}$73.33_{\pm18.86}$ & \cellcolor{orange!15}$90.47_{\pm2.92}$ \\
                \textit{improvement} & \textcolor{red}{$+8.71 \uparrow$} & \textcolor{red}{$+12.88 \uparrow$} & \textcolor{blue}{$-6.67 \downarrow$} & \textcolor{red}{$+6.94 \uparrow$} \\
                \arrayrulecolor{gray} \hline \arrayrulecolor{black}
                w/o Color & $86.95_{\pm6.15}$ & $90.74_{\pm13.09}$ & $73.33_{\pm18.86}$ & $91.11_{\pm5.12}$ \\
                w/ \hspace{1mm} Color & $91.48_{\pm5.76}$ & $94.74_{\pm7.44}$ & $80.00_{\pm0.00}$ & $94.44_{\pm3.93}$ \\
                \textit{improvement} & \textcolor{red}{$+4.53 \uparrow$} & \textcolor{red}{$+4.00 \uparrow$} & \textcolor{red}{$+6.67 \uparrow$} & \textcolor{red}{$+3.33 \uparrow$} \\
                \bottomrule
            \end{tabular}}
            \caption{Attribute Importance}
            \label{tab:sub5}
        \end{subtable} &&
        
        \begin{subtable}{0.4\textwidth}
            \captionsetup{font=normalsize}
            \small
            \renewcommand{\arraystretch}{1.1}
            \setlength{\tabcolsep}{0.5mm} 
            \centering
            \resizebox{\textwidth}{!}{
            \begin{tabular}{lcccc}
                \toprule
                \multirow{2}{*}{Attribute} & \multicolumn{4}{c}{\cellcolor{gray!20}DraWritePD} \\
                & Accuracy & Sensitivity & Specificity & $\text{F}_{1}$-score  \\
                \hline
                 w/o Height & $85.19_{\pm2.62}$ & $65.08_{\pm5.94}$ & $97.22_{\pm3.93}$ & $76.26_{\pm5.00}$ \\
                w/ \hspace{1mm} Height & \cellcolor{orange!15}$90.74_{\pm2.62}$ & \cellcolor{orange!15}$80.16_{\pm6.25}$ & \cellcolor{orange!15}$97.22_{\pm3.93}$ & \cellcolor{orange!15}$86.32_{\pm4.23}$ \\
                \textit{improvement} & \textcolor{red}{$+5.55 \uparrow$} & \textcolor{red}{$+15.08 \uparrow$} & \textcolor{gray}{$+0.00 =$} & \textcolor{red}{$+10.06 \uparrow$} \\
                \arrayrulecolor{gray} \hline \arrayrulecolor{black}
                w/o Color & $88.89_{\pm4.54}$ & $74.60_{\pm8.09}$ & $97.22_{\pm3.93}$ & $82.79_{\pm8.00}$ \\
                w/ \hspace{1mm} Color & $87.04_{\pm6.93}$ & $80.16_{\pm6.25}$ & $91.67_{\pm11.79}$ & $82.36_{\pm8.55}$ \\
                \textit{improvement} & \textcolor{blue}{$-1.85 \downarrow$} & \textcolor{red}{$+5.56 \uparrow$} & \textcolor{blue}{$-5.55 \downarrow$} & \textcolor{blue}{$-0.43 \downarrow$} \\
                \bottomrule
            \end{tabular}}
            \caption{Attribute Importance}
            \label{tab:sub6}
        \end{subtable} \\
    \end{tabular}}
    
\end{table*}

\subsubsection{Settings.} \label{sec:setting}
We implement PointExplainer using the Pytorch library \cite{paszke2019pytorch}, based on the model and code provided by \cite{qi2017pointnet, ribeiro2016should}. We train PointExplainer using CrossEntropy loss, AdamW optimizer \cite{kingma2014adam}, an initial learning rate $lr=1e-4$, weight decay $10^{-4}$, with Cosine Decay, and a batch size of $16$ for $300$ epochs. We perform stratified three-fold cross-validation at the individual level and report diagnostic performance in terms of accuracy, sensitivity, specificity, and $\text{F}_{1}$ score. Note that in our experiments, True positive (TP) represents the number of instances correctly predicted by the model as PD. For the SST, DST, and DraWritePD datasets, we use pressure, radius, and acceleration as the height attributes, respectively. The window size $w$ is set to $256$, $512$, and $512$ for these datasets, and different step sizes $s$ are used to maintain a roughly balanced number of segmented patches for each category (HC and PD), with a total of approximately $6,000$ patches. Additionally, to ensure the simplicity of the attribution map, we set the number of superpoints $M_{s}$ per stroke to $11$ and the number $K$ of perturbed samples to $200$. Unless otherwise stated, all experimental results are based on the decision threshold $\alpha=0.5$, and the point cloud representations only contain the three basic attributes $(x,y,z)$. 

\subsection{Diagnosis Robustness}
Table \ref{Tab.1} reports the diagnostic results of PointExplainer and other existing methods. (i) Diagnostic performance: PointExplainer presents competitive diagnostic results with robust performance, for example, with accuracy to $88.10\%$, $85.62\%$, and $90.74\%$, respectively. It clearly suggests the effectiveness of our proposed method. (ii) Representation method: We use the sparse 3D point cloud representation (e.g., only fusing three hand-drawn features), and even achieve superior diagnostic performance compared to other modality methods. This superiority illustrates that our representation method models more intuitive and efficient spatial hand-drawn patterns (as stated in Section \ref{sec:modelling}). (iii) Training process: Fig.\ref{fig:4} shows the stable convergence characteristics of PointExplainer. Combined with the test results in Table \ref{Tab.1}, the risk of model overfitting can be eliminated. (iv) Threshold adjustment: By optimizing the decision threshold $\alpha$, the diagnostic performance of PointExplainer can be further improved (see the last row of Table \ref{Tab.1} and Fig.\ref{fig:4}), for example, accuracy with an average improvement of $1.85\%$. Note that PointExplainer uses the most basic PointNet \cite{qi2017pointnet} as the recognition model and has strong potential for further improvement \cite{qi2017pointnet++,shen2018mining,wu2019pointconv,zhao2021point}. 

\begin{table*}[t]
    \caption{Ablation study of \textit{explanation robustness} on the SST, DST, and DraWritePD datasets. We additionally design a new perturbation strategy (as described in Section \ref{sec:explanation robustness}). The results indicate that under this new strategy, PointExplainer is still able to effectively provide reliable explanations, highlighting its robustness (see Section \ref{sec:perturnation} for details on the orginal perturbation strategy). Default settings are marked in\colorbox{orange!15}{orange}.}
    \label{Tab.4}
    \resizebox{\textwidth}{!}{
    \begin{tabular}{ccccc}
        \begin{subtable}{0.4\textwidth}
            \captionsetup{font=normalsize}
            \small
            \renewcommand{\arraystretch}{1.2}
            \setlength{\tabcolsep}{0.5mm} 
            \centering
            \resizebox{\textwidth}{!}{
            \begin{tabular}{lcccc}
                \toprule
                \multirow{2}{*}{Explainer} & \multicolumn{4}{c}{\cellcolor{gray!20}SST} \\
                &  PC $\downarrow$ & CA $\uparrow$ & AC $\uparrow$ & DA $\uparrow$ \\
                \hline
                LR  & \underline{$0.0244_{\pm0.0236}$} & \underline{$0.9868_{\pm0.1140}$} & \underline{$0.9120_{\pm0.1285}$} &  $0.9602_{\pm0.0694}$ \\
                Ridge & $0.0295_{\pm0.0272}$ &  \underline{$0.9868_{\pm0.1140}$} & $0.9119_{\pm0.1286}$ &  $0.9577_{\pm0.0697}$ \\
                ElasticNet & $0.0302_{\pm0.0264}$ &  \underline{$0.9868_{\pm0.1140}$} &  $0.9107_{\pm0.1306}$ & $\boldsymbol{0.9782_{\pm0.0516}}$ \\
                DT & $0.0416_{\pm0.0497}$ &  $0.9737_{\pm0.1601}$ & $0.8883_{\pm0.1204}$ &  $0.9672_{\pm0.0575}$ \\
                RF & $0.0397_{\pm0.0465}$ & \underline{$0.9868_{\pm0.1140}$} & $0.9080_{\pm0.0952}$ &  $0.9577_{\pm0.0665}$ \\
                XGB & \cellcolor{orange!15}$\boldsymbol{0.0150_{\pm0.0169}}$ & \cellcolor{orange!15}$\boldsymbol{1.0000_{\pm0.0000}}$ & \cellcolor{orange!15}$\boldsymbol{0.9423_{\pm0.0747}}$ &  \cellcolor{orange!15}\underline{$0.9697_{\pm0.0568}$} \\
                \bottomrule
            \end{tabular}}
            \caption{Explanation Robustness}
            \label{tab:sub1}
        \end{subtable}  &&
        
        \begin{subtable}{0.4\textwidth}
            \captionsetup{font=normalsize}
            \small
            \renewcommand{\arraystretch}{1.2}
            \setlength{\tabcolsep}{0.5mm} 
            \centering
            \resizebox{\textwidth}{!}{
            \begin{tabular}{lcccc}
                \toprule
                \multirow{2}{*}{Explainer} & \multicolumn{4}{c}{\cellcolor{gray!20}DST} \\
                & PC $\downarrow$ & CA $\uparrow$ & AC $\uparrow$ & DA $\uparrow$  \\
                \hline
                 LR  & $0.0214_{\pm0.0299}$ & \underline{$0.9855_{\pm0.1195}$} & $0.7474_{\pm0.2675}$ & $0.9179_{\pm0.0921}$ \\
                Ridge & $0.0214_{\pm0.0225}$ &  \underline{$0.9855_{\pm0.1195}$} & $0.7469_{\pm0.2679}$ & $0.9179_{\pm0.0921}$ \\
                ElasticNet & $0.0209_{\pm0.0225}$ &  \underline{$0.9855_{\pm0.1195}$} & $0.7493_{\pm0.2659}$ & $0.9271_{\pm0.0914}$ \\
                DT & $0.0181_{\pm0.0291}$ &  $0.9420_{\pm0.2337}$ & $0.8090_{\pm0.2164}$ & \underline{$0.9390_{\pm0.0774}$} \\
                RF & \underline{$0.0146_{\pm0.0218}$} &  $0.9565_{\pm0.2039}$ & 
                \underline{$0.8320_{\pm0.2098}$} & $0.9296_{\pm0.0940}$ \\
                XGB & \cellcolor{orange!15}$\boldsymbol{0.0094_{\pm0.0136}}$ & \cellcolor{orange!15}$\boldsymbol{1.0000_{\pm0.0000}}$ & \cellcolor{orange!15}$\boldsymbol{0.8690_{\pm0.1722}}$ & \cellcolor{orange!15}$\boldsymbol{0.9473_{\pm0.0716}}$ \\
                \bottomrule
            \end{tabular}}
            \caption{Explanation Robustness}
            \label{tab:sub2}
        \end{subtable}  &&
        
        \begin{subtable}{0.4\textwidth}
            \captionsetup{font=normalsize}
            \small
            \renewcommand{\arraystretch}{1.2}
            \setlength{\tabcolsep}{0.5mm} 
            \centering
            \resizebox{\textwidth}{!}{
            \begin{tabular}{lcccc}
                \toprule
                \multirow{2}{*}{Explainer} & \multicolumn{4}{c}{\cellcolor{gray!20}DraWritePD} \\
                & PC $\downarrow$ & CA $\uparrow$ & AC $\uparrow$ & DA $\uparrow$  \\
                \hline
                LR  & $0.1205_{\pm0.0739}$ &  \underline{$0.9815_{\pm0.1348}$} & $0.5839_{\pm0.3109}$ & $0.8552_{\pm0.1694}$ \\
                Ridge & $0.1187_{\pm0.0756}$ &  \underline{$0.9815_{\pm0.1348}$} & $0.5840_{\pm0.3107}$ & $0.8519_{\pm0.1685}$ \\
                ElasticNet & $0.1201_{\pm0.0752}$ & \underline{$0.9815_{\pm0.1348}$} & $0.5843_{\pm0.3100}$ & \underline{$0.8623_{\pm0.1643}$} \\
                DT & \underline{$0.0771_{\pm0.0710}$} &  $\boldsymbol{1.0000_{\pm0.0000}}$ & \underline{$0.6538_{\pm0.2522}$} &  $0.8460_{\pm0.1569}$ \\
                RF & $0.0790_{\pm0.0718}$ & $\boldsymbol{1.0000_{\pm0.0000}}$ & $0.6534_{\pm0.2901}$ & $0.8590_{\pm0.1501}$ \\
                XGB & \cellcolor{orange!15}$\boldsymbol{0.0440_{\pm0.0392}}$ & \cellcolor{orange!15}$\boldsymbol{1.0000_{\pm0.0000}}$ & \cellcolor{orange!15}$\boldsymbol{0.7109_{\pm0.2928}}$ & \cellcolor{orange!15}$\boldsymbol{0.8859_{\pm0.1417}}$ \\
                \bottomrule
            \end{tabular}}
            \caption{Explanation Robustness}
            \label{tab:sub3}
        \end{subtable}  \\
        \end{tabular}}
    
\end{table*}

\begin{figure*}[t]
  \centering  
  \vspace{-1mm}
  \includegraphics[width=1.0\textwidth]{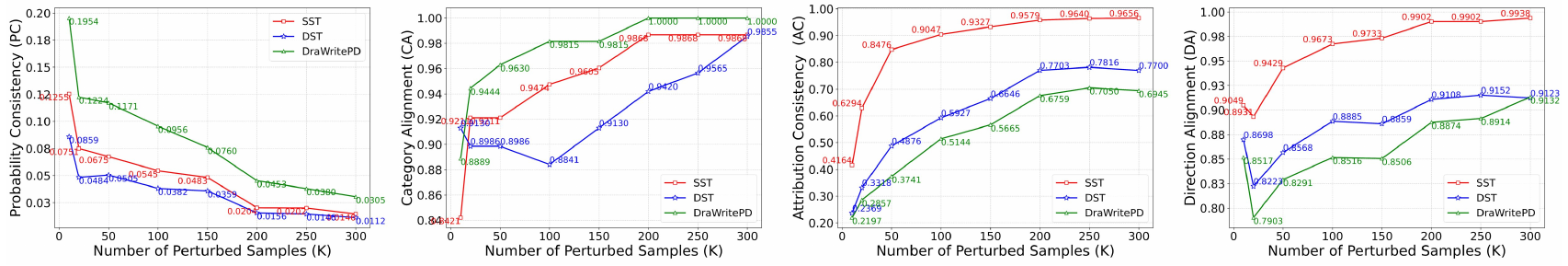}
  \vspace{-1mm}
  \caption{Ablation study of \textit{the number of perturbed samples} $K$ on the SST, DST, and DraWritePD datasets. We set $K=200$ by default in our experiments.}
  \label{fig:6}
  \vspace{-3mm}
\end{figure*}

\subsection{Explanation Faithfulness}
Table \ref{Tab.2} presents the benchmark results of PointExplainer's explanation faithfulness. (i) Quantitative evaluation: The metric results, including PC, CA, AC, and DA (as defined in Section \ref{sec:verify}), demonstrate that PointExplainer can effectively locate the key hand-drawn segments that drive the model’s diagnostic decisions. In addition, PointExplainer performs robustly when coupled with both linear and nonlinear interpretable models. As expected, using the XGBoost regressor (XGB) as the explainer $G$ achieves the best explanation quality on 11/12 metrics (4 metrics in 3 datasets). (ii) Qualitative analysis: We further evaluate PointExplainer's explanations by selectively masking the located key superpoints and observing the changes in diagnostic decisions. Specifically, Fig.\ref{fig:5} shows the attribution maps and the perturbation results based on these maps. The results indicate that masking the located superpoints basically leads to the decision changes (increase or decrease) of the corresponding attributes (positive or negative). In particular, the changes are more significant when masking key superpoints (i.e., with large absolute attribution values), although masking non-key superpoints (i.e., with small absolute attribution values) may occasionally make mistakes. This demonstrates that our PointExplainer successfully identifies important hand-drawn regions. (iii) Case explanation: Taking Fig.\ref{fig:5} (i) as an example, the key superpoint (i.e., the dark red region) identified by PointExplainer is the hand-drawn breakpoint pattern, a plausible feature that is more commonly observed in PD patients. This finding indicates some instances processed by our method even concur with human intuition well, further intuitively supports the reliability of our PointExplainer's explanation.

\subsection{Ablation Study}
We ablate our PointExplainer using the default settings as described in Section \ref{sec:setting}. Several intriguing properties are observed.

\noindent\textbf{Height Feature.} 
We compare the diagnostic differences using different hand-drawn features as the point cloud height in Table \ref{Tab.3} (a-c). Azimuth, altitude, and pressure are recorded features, and radius, velocity, and acceleration are derived features. The results show that while the optimal choice varies across datasets, using radius as the height attribute consistently performs well, for example, with accuracy reaching $88.21\%$, $85.62\%$, and $85.19\%$, respectively. In addition, the selection of height attribute significantly impacts diagnostic performance, with accuracy gaps up to $\sim10\%$ ($76.31\%$ $vs.$ $88.21\%$, $70.98\%$ $vs.$ $85.62\%$, $79.63\%$ $vs.$ $90.74\%$), highlighting its importance in our representation method. Other hand-drawn features could also be considered as potential candidates, but we do not discuss them further here.

\begin{figure*}[t]
  \centering  
  \vspace{-1mm}
  \includegraphics[width=1.0\textwidth]{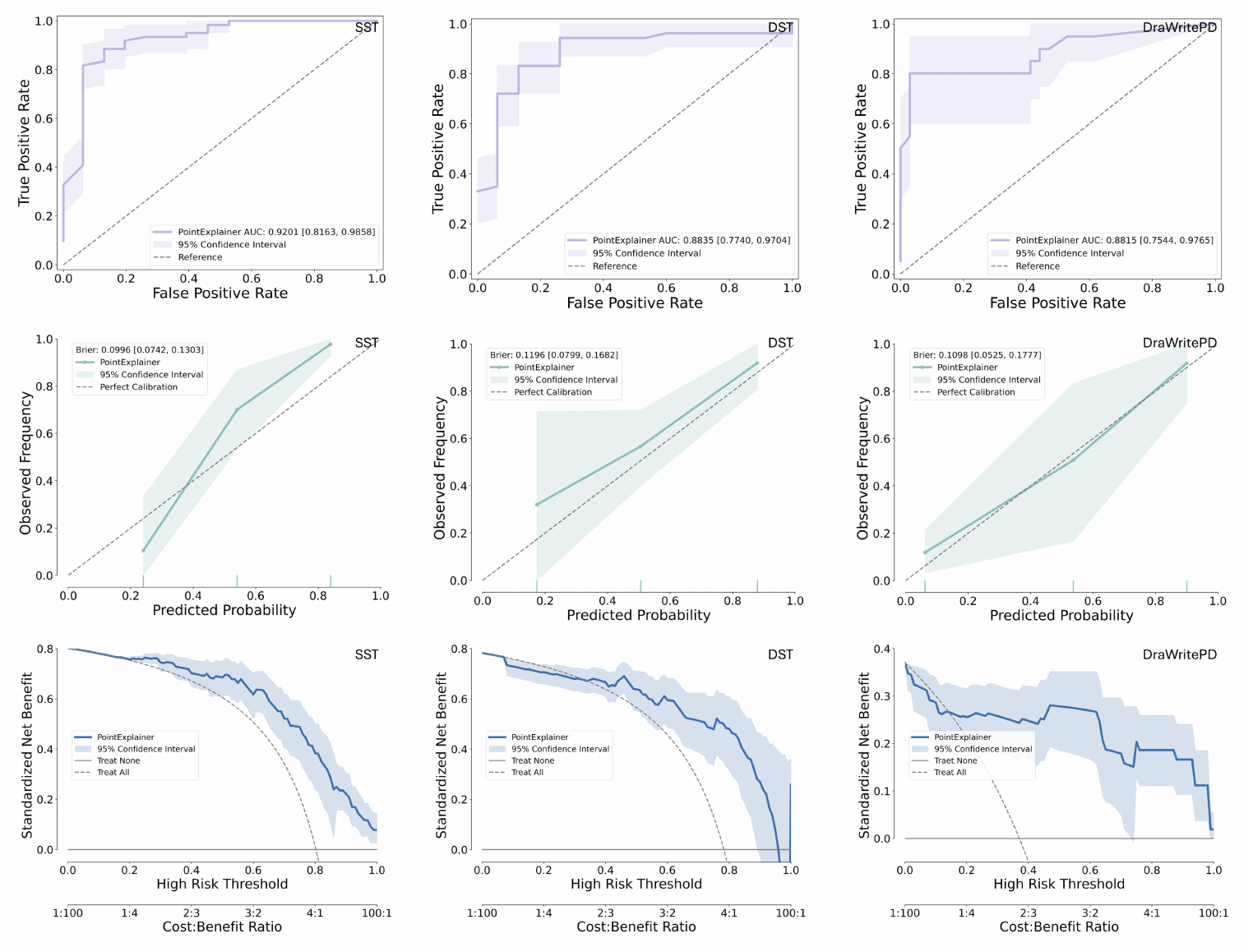}
  \vspace{-1mm}
  \caption{\textbf{Potential Clinical Applicability of PointExplainer}: (i) Discrimination, assessed using the Receiver Operating Characteristic (ROC) curve (first row); (ii) Calibration, evaluated via the calibration curve (second row); and (iii) Clinical Benefit, analyzed through Decision Curve Analysis (DCA) (third row). Mean curves are computed from 1,000 bootstrap resamples with replacement, with shaded areas representing the 95$\%$ confidence intervals.}
  \label{fig:7}
\end{figure*}

\noindent\textbf{Window Size.}
PointExplainer is a local feature-based method and its performance is influenced by the spatial semantic structure of the patch (as stated in Section \ref{sec:diagnosis}), which makes the window size $w$ an important parameter. In Table \ref{Tab.3} (d-f), we evaluate six different window sizes ranging from $64$ to $1024$. The results exhibit a consistent trend: for example, accuracy initially improves steadily with increasing window size before declining, reaching optimal values at $256$, $512$, and $512$, respectively. In addition, compared to the selection of height attributes, diagnostic performance is less sensitive to window size, particularly on the DraWritePD dataset, with an accuracy fluctuation of only $\sim5\%$ ($85.19\%$–$90.74\%$). Notably, we did not further fine-tune the window size, but it is expected that appropriate adjustments could bring slight performance improvements.

\noindent\textbf{Attribute Importance.}
We investigate the role of different attributes (i.e., height and color) in point cloud representation (as stated in Section \ref{sec:modelling}) and design a fair comparative experiment. Our results thus far are based on point clouds with three attributes $(x, y, z)$. In Table \ref{Tab.3} (g-i), the experiments are divided into two independent parts: in the Height experiment, we use the default three attributes $(x, y, z)$ to represent the point cloud, with the \textit{w/o Height} setting fixing $z = 1$; in the Color experiment, we use six attributes $(x, y, z, r, g, b)$ to represent the point cloud, with the \textit{w/o Color} setting fixing $(r, g, b) = (1,1,1)$. The results show that incorporating the height attribute leads to significant improvements on 9/12 metrics (4 metrics in 3 datasets), for example, with accuracy increasing by $16.87\%$, $8.71\%$, and $5.55\%$, respectively. However, the introduction of color attributes does not yield the expected gains and instead results in performance fluctuations, for example, with accuracy decreasing by $2.66\%$, increasing by $4.53\%$, and decreasing by $1.85\%$, respectively. This indicates that certain additional attributes, such as color, may not be essential within our representation method. Notably, compared to other related works (e.g., \cite{drotar2014analysis,impedovo2019velocity,diaz2021sequence,wang2024lstm}) that rely on more hand-drawn features, our default point cloud representation method (i.e., the \textit{w/ Height} setting in Table \ref{Tab.3} (g-i)) still achieves competitive diagnostic performance.

\noindent\textbf{Explanation Robustness.}\label{sec:explanation robustness}
We investigate the stability of PointExplainer's explanation ability by varying the perturbation strategy. Section \ref{sec:perturnation} describes the default perturbation strategy, with the corresponding experimental results shown in Table \ref{Tab.2}. Here, we additionally design a new perturbation strategy: keeping the 2D coordinates $(x,y)$ of points unchanged while only setting the height attribute $z$ to a constant value (e.g., the average height of all points in the superpoint). We assume that this eliminates the spatial semantic structure information provided by the height attribute. Table \ref{Tab.4} studies this new strategy. The results show that PointExplainer continues to deliver strong performance metrics, especially when coupled with XGB, where it consistently achieves the best explanation quality on 11/12 metrics (4 metrics in 3 datasets). This indicates that PointExplainer retains its explanation quality even under modified perturbation conditions, emphasizing its robustness in different perturbation strategies.

\noindent\textbf{Number of Perturbed Samples.}
The perturbed samples are used to train the explainer $G$, enabling it to approximate the decision behavior of the diagnostic model $F$ (as stated in Section \ref{sec:explain}). The number $K$ of perturbed samples is important to the quality of the explanation. An appropriate number of perturbed samples can ensure stable explanations while minimizing training costs. Fig.\ref{fig:6} shows the impact of the number of perturbed samples. As expected, as the number of samples increases, the fit of the explainer to the diagnostic model gradually improves and tends to saturation. We set $K = 200$ by default, which performs well on different metrics and datasets.

\subsection{Clinical Applicability}
We evaluate the potential clinical applicability \cite{alba2017discrimination, steyerberg2010assessing,subramaniam2024grand} of our PointExplainer in Fig. \ref{fig:7}.

\noindent\textbf{Discrimination} \cite{alba2017discrimination} measures the model's ability to correctly differentiate between categories, such as PD patients $vs.$ HC individuals. As shown in the first row of Fig.\ref{fig:7}, we use the Area Under the Receiver Operating Characteristic Curve (AUC-ROC) \cite{chunhabundit2025sex} as the metric. AUC=1.0 (i.e., the ROC curve close to the upper right corner (0,1)) indicates perfect classification, and AUC=0.5 (i.e., the ROC curve close to the reference line) is equivalent to random guessing. Therefore, a higher AUC score demonstrates a model's superior ability to correctly classify samples. Our PointExplainer achieves AUC scores of $0.9201$, $0.8835$, and $0.8815$ respectively, indicating the well discrimination ability.

\noindent\textbf{Calibration} \cite{boehm2025multimodal} measures whether the predicted probabilities of the model align with the actual incidence of the disease, preventing the model from over-intervention or risk underestimation. As shown in the second row of Fig.\ref{fig:7}, we use the calibration curve \cite{huang2020tutorial} and the Brier score \cite{steyerberg2010assessing} as metrics. Ideally, the calibration curve should be as close to the diagonal $y=x$ as possible, indicating a strong agreement between predicted probabilities and observed incidence rates. If the curve is above the diagonal, it indicates that the model is overconfident and may result in unnecessary medical costs; and if the curve is below the diagonal, it indicates that the model underestimates the risk of the disease and may cause patients to fail to receive timely treatment. A Brier score $<0.25$ \cite{ramirez2025systematic,butner2024hybridizing} is generally considered to indicate well calibration, and our PointExplainer obtains Brier scores of $0.0996$, $0.1196$, and $0.1098$ respectively. These low Brier scores indicate that PointExplainer performs well in calibration.

\noindent\textbf{Clinical Benefit} \cite{zhang2024retfound} measures whether the model contributes to improving clinical decision-making in real-world medical applications. As shown in the third row of Fig.\ref{fig:7}, we use decision curve analysis (DCA) \cite{zhang2024retfound} for evaluation, where the x-axis represents the decision threshold $\alpha$, and the y-axis indicates the net benefit at that threshold, which quantifies the trade-off between true positives and false positives. The curves ``Treat None'' and ``Treat All'' represent two extreme strategies, i.e., assuming all subjects are negative or positive, respectively, as baseline references for comparison. The results show that whether under the default threshold of $0.5$ or under the optimal threshold ($0.52$, $0.46$, $0.50$, respectively, see Fig.\ref{fig:4}), the benefit curves consistently outperform these two reference lines, indicating that our PointExplainer can bring positive clinical decision-making benefits.

\section{Discussion}
Reliable diagnostic strategies that interpret well are essential for medical artificial intelligence \cite{carusi2023medical,hillis2024health}. In analyzing digitized hand-drawn signals for PD diagnosis, while existing works have made progress \cite{aouraghe2022literature}, most designs (e.g., \cite{drotar2014analysis,impedovo2019velocity,diaz2021sequence,wang2024lstm}) are still performance-oriented. Enhancing the interpretability within this community has the potential to foster collaboration between clinicians and algorithms, and build greater trust. In this study, we reconstruct hand-drawn trajectories and observe differences (e.g., smooth and jittery) in local hand-drawn regions, in turn proposing PointExplainer to investigate the feasibility of assigning attribution values to hand-drawn segments. As an explainable diagnostic strategy, PointExplainer presents the model decision logic in the form of intuitive attribution maps, and shows robust performance and reliable explanation in systematic evaluations. Notably, our additional foray into the potential clinical applicability is exploratory and we believe it can be improved with more effort, but regardless we expect the perspective of this work.

\section{Conclusion}
This paper presents PointExplainer, an intuitive and explainable strategy for early diagnosis of PD. It constructs an attribution map to measure the contribution of local hand-drawn regions in model decisions. Extensive experiments and ablations comprehensively evaluate the effectiveness and robustness of our method. Future work may focus on investigating whether typical and intuitive hand-drawn patterns exist within the identified regions, aiming to establish their correspondence to clinical significance. It is possible to be addressed with an unsupervised clustering strategy, which is left for future work.

\section*{Acknowledgments}
This work was supported by the Estonian Research Council grant PRG $2100$. It was also partially supported by the FWO Odysseus 1 grant G.0H94.18N: Analysis and Partial Differential Equations, and the Methusalem programme of the Ghent University Special Research Fund (BOF) (Grant number 01M01021). Michael Ruzhansky is also supported by EPSRC grant EP/R003025/2.

\bibliographystyle{plain}
\bibliography{refs}

\end{document}